\documentclass[runningheads]{llncs}

\usepackage{eccv}

\usepackage{eccvabbrv}

\usepackage{graphicx}
\usepackage{booktabs}
\usepackage{amsmath}
\usepackage{dsfont}
\usepackage{color}
\usepackage[ruled,vlined,linesnumbered]{algorithm2e}

\usepackage{xcolor}

\SetCommentSty{mycommfont}

\newcommand{\acronym}{COIN\xspace}

\usepackage[accsupp]{axessibility}  

\usepackage[pagebackref,breaklinks,colorlinks,citecolor=eccvblue]{hyperref}

\usepackage{orcidlink}

\begin{document}


\title{\acronym: Control-Inpainting Diffusion Prior for Human and Camera Motion Estimation\vspace{-5mm}}

\titlerunning{\acronym: Control-Inpainting Diffusion}

\author{Jiefeng Li$^{1,2}$ \and
Ye Yuan$^1$ \and
Davis Rempe$^1$ \and
Haotian Zhang$^1$ \and
Pavlo Molchanov$^1$ \and
Cewu Lu$^2$ \and
Jan Kautz$^1$ \and
Umar Iqbal$^1$
}

\authorrunning{L.~Jiefeng et al.}

\institute{$^1$NVIDIA \quad$^2$Shanghai Jiao Tong University
\\
}

\maketitle

\begin{center}
    \vspace{-6mm}
    \href{https://nvlabs.github.io/COIN/}{https://nvlabs.github.io/COIN/}
\end{center}

\setlength{\textfloatsep}{0pt}
\begin{center}
    \captionsetup{type=figure}
    \includegraphics[width=\linewidth]{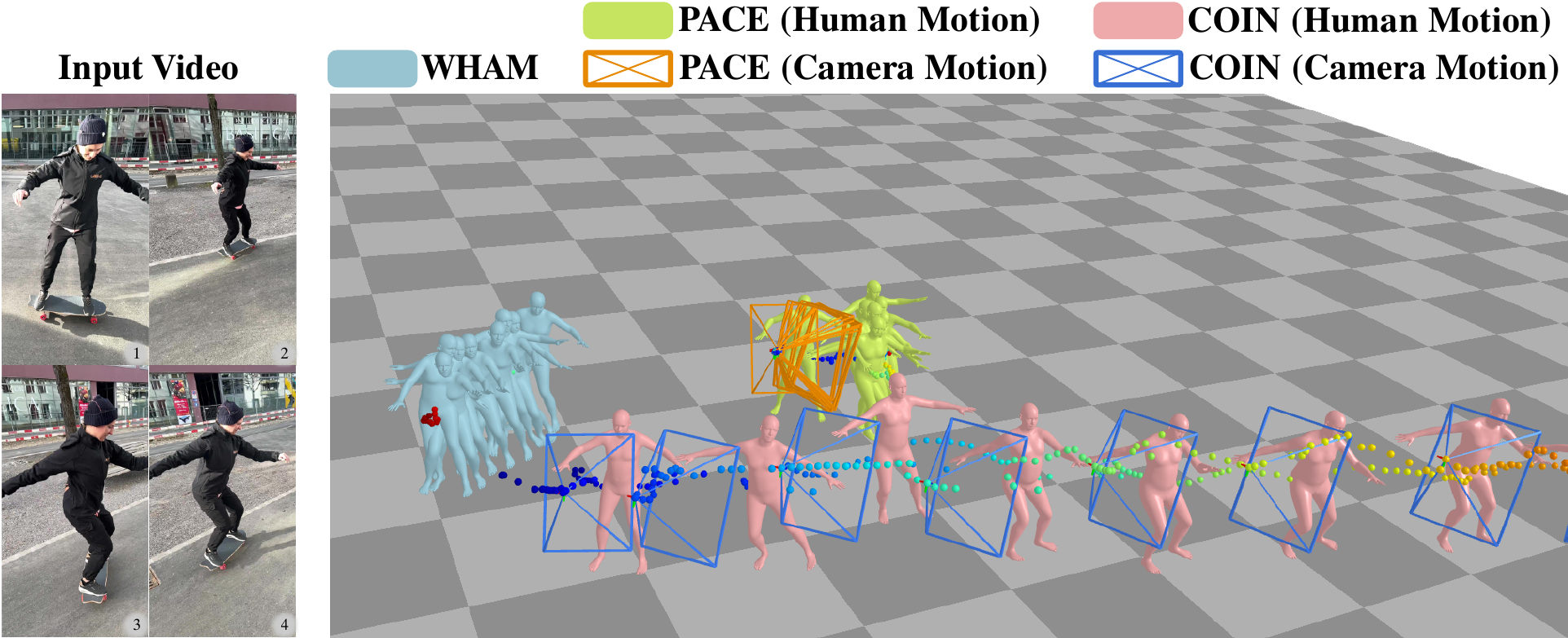}
    \vspace{-6mm}
    \caption{Capturing global human and camera motion from a dynamic camera presents unique challenges. In the input video, a person is riding a skateboard – while the local body motion may remain relatively constant, the global position of the individual changes significantly. Current state-of-the-art methods such as PACE~\cite{kocabas2024pace} and WHAM~\cite{shin2024wham} fail catastrophically on such out-of-distribution motions. Our approach, COIN, gracefully handles such challenging cases, owing to our control-inpainting motion diffusion prior and novel human-scene relation loss.}
    \label{fig:teaser}
\end{center}

\begin{abstract}
\vspace{-1mm}

  Estimating global human motion from moving cameras is challenging due to the entanglement of human and camera motions. To mitigate the ambiguity, existing methods leverage learned human motion priors, which however often result in oversmoothed motions with misaligned 2D projections. To tackle this problem, we propose \acronym, a control-inpainting motion diffusion prior that enables fine-grained control to disentangle human and camera motions. Although pre-trained motion diffusion models encode rich motion priors, we find it non-trivial to leverage such knowledge to guide global motion estimation from RGB videos. \acronym introduces a novel control-inpainting score distillation sampling method to ensure \textit{well-aligned}, \textit{consistent}, and \textit{high-quality} motion from the diffusion prior within a joint optimization framework. Furthermore, we introduce a new human-scene relation loss to alleviate the scale ambiguity by enforcing consistency among the humans, camera, and scene. Experiments on three challenging benchmarks demonstrate the effectiveness of \acronym, which outperforms the state-of-the-art methods in terms of global human motion estimation and camera motion estimation. As an illustrative example, COIN outperforms the state-of-the-art method by 33\% in world joint position error (W-MPJPE) on the RICH dataset.

\end{abstract}

\section{Introduction}
\label{sec:intro}

Recovering \emph{global} human and camera motion from dynamic RGB videos is an important problem with many applications, such as animation, human-computer interaction, mixed reality, and robotics. However, it is a very challenging problem due to the entanglement of human and camera motion. 

There are only a few works~\cite{yuan2022glamr, ye2023slahmr, kocabas2024pace, shin2024wham}  that try to address this problem. Earlier methods~\cite{yuan2022glamr,li2022dnd} only focus on human motion and ignore the camera motion. Their core insight is that the global body motion is highly correlated with the local motion. Thus, they can use the local body movements to estimate the global orientation and trajectory with a regression model~\cite{yuan2022glamr} or by combining them with physics constraints~\cite{li2022dnd}. However, these regression models ignore the camera movements, so they fail to maintain consistency with the input video, whereas physics-based methods fail to model complex in-the-wild environments so are limited to controlled scenarios. Recent works~\cite{ye2023slahmr,kocabas2024pace} try to jointly estimate the human and camera motion by exploiting learned motion priors~\cite{rempe2021humor, he2022nemf} and SLAM~\cite{opensfm2021, teed2021droid, teed2023deep}. They try to constrain the human body motion in a low-dimensional latent space of a motion prior model, which results in reconstructed motions that are overly smooth and do not align well with video observations. Moreover, the optimization of the camera motion is only based on the global human motion from the motion prior. Hence, they fail catastrophically if the initial human motion predictions are significantly incorrect (as shown in Fig.~\ref{fig:teaser}).

More recently, Denoising Diffusion Models~\cite{ho2020denoising, tevet2022human} have emerged as a powerful family of generative models that can model high-quality data priors. Nonetheless, effectively leveraging the learned priors remains an ongoing challenge. Score Distillation Sampling (SDS) is commonly employed for this purpose~\cite{poole2023dreamfusion}, but we find that naive application of SDS also results in inconsistencies with the available observations (see \cref{sec:ablation}). The root cause of this problem lies in the inconsistency of randomly sampled motions during SDS optimization. Without constraints, these motions may not align with observed evidence, leading to overly smoothed results that lack detail due to the mode-averaging effect.

In this work, we propose \acronym, a hybrid \textbf{CO}ntrol-\textbf{IN}painting score distillation sampling method to address the aforementioned limitations of vanilla SDS. 
First, we use the partially observed human motion from the video as \textit{control} signals to guide motion sampling. 
To tackle noisy observations which may be out of distribution for the prior, we propose a dynamic controlled sampling technique that iteratively refines the observed motions and updates the control signals to ensure effective distillation from the motion prior. Second, to further improve the consistency of the sampled motions, we also develop a novel \textit{soft inpainting} strategy. We automatically identify the high-confidence regions of the initial predicted global motion from the video and use them as soft constraints during optimization. Concretely, we sample less confident regions from scratch using the motion model, while the confident regions are only slightly refined. This ensures that the reconstructed motions do not deviate from the available observations. 
Our new SDS formulation is used to jointly optimize the human and camera motion by finding the most plausible solution that explains the observed evidence. Finally, to prevent catastrophic failure in cases where the initial body motion fails significantly,   
we propose a human-scene relation loss to consider the human-scene depth relations. This novel loss provides complementary information to the human motion prior by using local motion and scene features.
It regularizes the camera scale by enforcing consistency among the human motion, camera motion, and scene features.

We benchmark our approach on the synthetic HCM~\cite{kocabas2024pace} dataset and the real-world RICH~\cite{huang2022cap} and EMDB~\cite{kaufmann2023emdb} datasets. We demonstrate that our approach significantly outperforms the state-of-the-art methods in terms of human motion estimation and camera motion estimation.
Overall, the contributions of this paper can be summarized as follows:
\begin{itemize}
    \item We propose a novel control-inpainting motion prior specifically designed for global human motion estimation, which enhances score distillation sampling with dynamic control and soft inpainting to reconstruct well-aligned, consistent, and high-quality motions from video observations.
    \item We propose a new human-scene relation loss to resolve the scale ambiguity of the camera motion by enforcing consistency among the human motion, camera motion, and scene features.
    \item Our approach significantly outperforms the state-of-the-art methods in terms of human motion estimation and camera motion estimation on both synthetic and real-world datasets. In terms of global human motion estimation in world space, we outperform the state-of-the-art method PACE~\cite{kocabas2024pace} by $44\%$ and $33\%$ on the HCM~\cite{kocabas2024pace} and RICH~\cite{huang2022cap} datasets, respectively. We also compare with the contemporary work WHAM~\cite{shin2024wham}  and outperform it by $49\%$ and $7\%$ on the RICH and EMDB datasets, respectively.
\end{itemize}

\section{Related Work}
\label{sec:related}

\subsubsection{Camera-Space Human Pose Estimation.} Most existing works focus on root-relative local human pose estimation to bypass the difficulty in monocular depth estimation~\cite{Akhter:CVPR:2015, bogo2016keep, lassner2017unite, hmrKanazawa18, pavlakos2018humanshape, guler2019holo, kolotouros2019spin, pavlakos2019texture, Rong_2019_ICCV, kolotouros2019convolutional, choutas2020expose, zanfir2020weakly, sun2019human, joo2021eft, choi2020pose, kundu2020mesh, SMPL-X:2019, xu2019denserac, monototalcapture2019, song2020human, zhang2020object, zhou2021monocular, moon2020i2l, lin2021end, Mueller:CVPR:21, kolotouros2021prohmr, Zhang_2021_ICCV, Sun_2021_ICCV, humanMotionKanazawa19, kocabas2020vibe, luo20203d, choi2020beyond, rempe2021humor}. These methods ignore the position of the person in the camera coordinates. To overcome this limitation, recent methods estimate camera-space human poses by regression~\cite{zanfir2018monocular, jiang2020coherent, moon2019camera, Zanfir_2021_ICCV, ICG, Zhang_2021_CVPR, Xie_2021_ICCV, PhysCapTOG2020, liu20204d, zanfir2018monocular, Weng_2021_CVPR, li2020hybrik, iqbal2021kama, reddy2021tesstrack} or optimization~\cite{mono20173dhp, mehta2017vnect, XNect_SIGGRAPH2020,zanfir2018deep, rogez2017lcr}. Physics-based constraints are widely used to ensure the plausibility of the estimated poses~\cite{PhysCapTOG2020, iqbal2020learning, Xie_2021_ICCV, GraviCap2021, yuan2021simpoe, isogawa2020optical, gartner2022trajectory}. In addition to direct regression, heatmap-based representations have also been used to predict the absolute depths of multiple people~\cite{Fabbri_2020_CVPR,zhen2020smap,Sun:CVPR:2022}. A few methods improve absolution depth estimation by using predicted camera parameters~\cite{Kocabas_SPEC_2021, Zanfir_2021_ICCV, li2022cliff} instead of the predefined focal length. Despite the promising results for camera-space pose estimation, how to decouple the camera movement and estimate global human motions is still an open problem.

\vspace{-4mm}
\subsubsection{Monocular Global Human Pose Estimation.}
Recovering the global human motion from a monocular moving camera is challenging due to the entanglement of human and camera motions. To disentangle the camera movement, several methods use IMU sensors or pre-scanned environments to recover global human motions~\cite{vonMarcard2018, hassan2019resolving, hps2021Vladmir, pavlakos2022one}, which is impractical for large-scale adoption. Recent works use human motion priors~\cite{yuan2022glamr} or physics-based constraints~\cite{luo2021dynamics,li2022dnd} to recover human motions from monocular videos, but do not consider background scene features, which limits performance on in-the-wild videos. Sun~\etal~\cite{sun2023trace} use optical flow as a motion cue to estimate the global motions. A contemporary work, WHAM~\cite{shin2024wham}, uses a lifting network to estimate global human motions from 2D keypoints and camera angular velocities. While these works can estimate accurate global human motions, they do not recover camera motions. To explicitly recover the camera motion, Liu~\etal~\cite{liu20204d} use SLAM and convert the local pose from the camera to global coordinates. BodySLAM~\cite{henning2022bodyslam} jointly optimizes the human and camera motion using features of both humans and scenes. Along this line, SLAHMR~\cite{ye2023slahmr} and PACE~\cite{kocabas2024pace} use SLAM to initialize camera motions and optimize the camera using human motion priors~\cite{rempe2021humor,he2022nemf}. However, these methods rely on the human motion priors to regularize the camera motion, which may lead to inaccurate camera motion when the human motion is not well initialized (as shown in Fig.~\ref{fig:teaser}). Such wrong camera trajectories will further affect the optimization of human motions. In contrast, our approach relies on the consistency among the local human motion, scene features, and the camera for optimization, which provides information that complements the human motion priors and enables accurate estimation of both human and camera motions.

\vspace{-4mm}
\subsubsection{Human Motion Priors.}
There are a significant amount of approaches proposed to study 3D human dynamics for motion prediction and synthesis\cite{fragkiadaki2015recurrent,jain2016structural,li2017auto,martinez2017human,villegas2017learning,pavllo2018quaternet,aksan2019structured,gopalakrishnan2019neural,yan2018mt,barsoum2018hp,yuan2019diverse,hernandez2019human,kaufmann2020convolutional,yuan2020dlow,harvey2020robust,yuan2020residual,cao2020long,petrovich2021action,khurana2021detecting,hassan2021stochastic}. These learned human motion priors are used to help resolve pose ambiguity~\cite{kocabas2020vibe,rempe2021humor,zhang2021learning,he2022nemf} in human pose estimation. Recently, diffusion models~\cite{sohl2015deep} have also been used as priors for motion synthesis and infilling~\cite{tevet2022human,zhang2022motiondiffuse,yuan2023physdiff,huang2023diffusion,karunratanakul2023optimizing}. RoHM~\cite{zhang2024rohm}, adopts motion diffusion model to recover human motions from noisy and occluded input data. Xie~\etal ~\cite{xie2023omnicontrol} use spatial control signals to guide motion generation. They focus on generating realistic human motion given clean spatial constraints.
Müller~\etal~\cite{muller2023generative} build a diffusion model to learn the joint distribution over the poses of two people. They use the SDS loss to guide the generation of static poses. In contrast, we focus on dynamic human motions. We find adopting SDS directly for temporal human motions encounters the inconsistency issue. Therefore, to distill knowledge from the motion diffusion model, we propose a novel control-inpainting SDS to generate high-quality and consistent motion that aligns with observed evidence.

\section{Method}
\label{sec:method}

\begin{figure}[t]
    \centering
    \includegraphics[width=\linewidth]{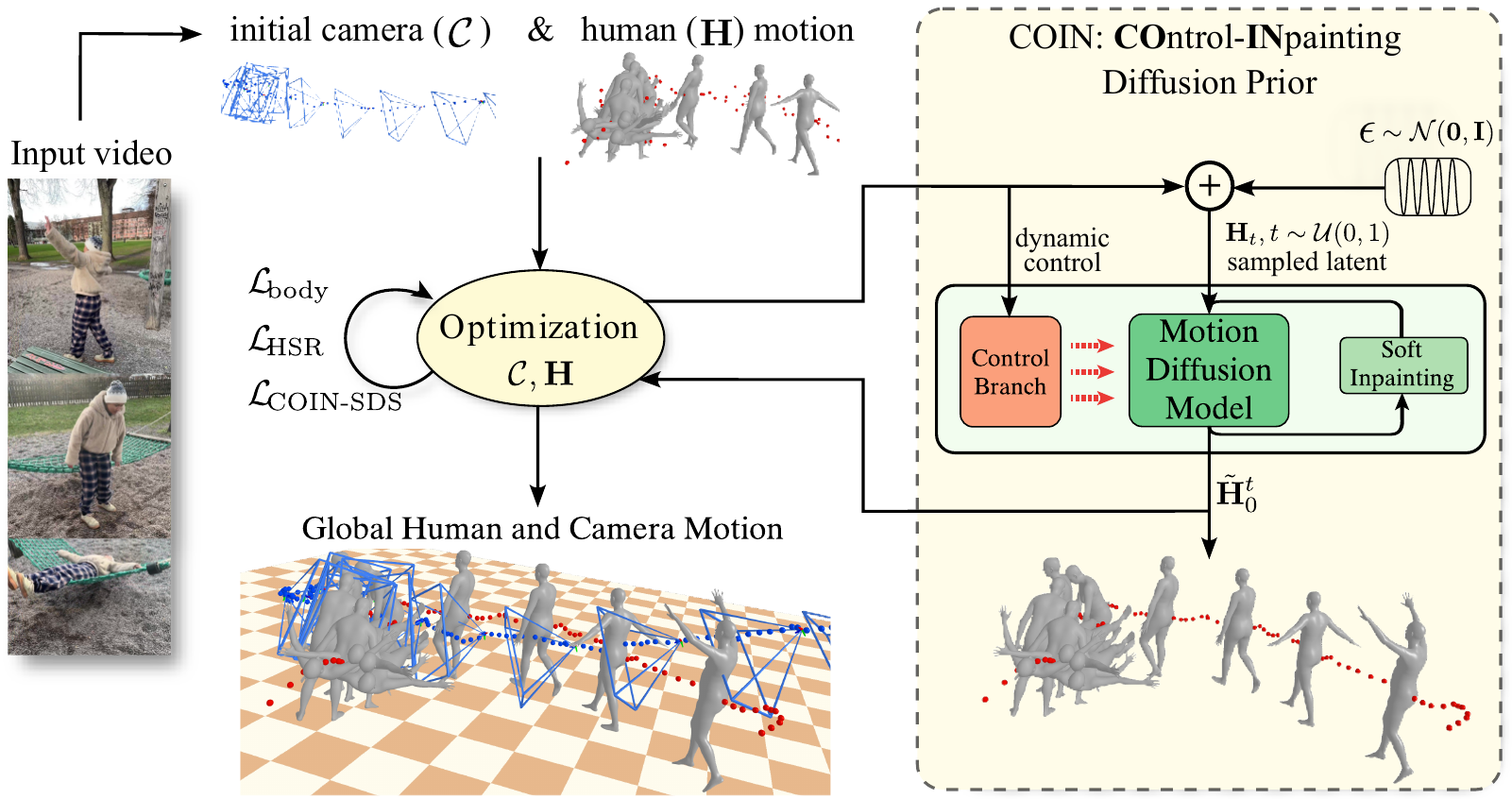}
    \vspace{-6mm}
    \caption{\textbf{Overview}. Given a video with a moving camera, we recover the global human motion $\mathbf{H}$ and camera motion $\mathcal{C}$ using an iterative optimization framework.
    We propose a novel Control-Inpainting SDS loss ($\mathcal{L}_\textrm{COIN-SDS}$) to leverage motion diffusion models as a prior. COIN-SDS is designed such that the sampled motions from the motion prior are consistent with video observations. We achieve this by controlling and constraining the sampling process of the motion diffusion model through novel control and soft-inpainting branches. We also propose a novel human-scene relation loss ($\mathcal{L}_\textrm{HSR}$) to encourage consistency among the human motion, camera motion, and scene features.  }
    \vspace{2mm}
    \label{fig:overview}
\end{figure}

The overall framework of \acronym is illustrated in Fig.~\ref{fig:overview}. Given an in-the-wild RGB video with $T$ frames captured by a dynamic camera, our goal is to estimate both the global human motion $\mathbf{H} = \{\mathbf{h}^{(1)}, \mathbf{h}^{(2)}, \ldots, \mathbf{h}^{(T)}\}$ and the camera motion $\mathcal{C} = \{\mathbf{c}^{(1)}, \mathbf{c}^{(2)}, \ldots, \mathbf{c}^{(T)}\}$ in a global world coordinate system.
We use off-the-shelf 3D human pose and shape estimation method HybrIK~\cite{li2020hybrik} to obtain per-frame initial SMPL parameters in the camera space and DROID-SLAM~\cite{teed2021droid} to obtain the initial per-frame camera-to-world transforms. We convert the local human motion to the world coordinates with the estimated camera. However, because the camera trajectories from SLAM are up to an unknown scale, the initial global human motion will abnormally drift and float in the world space. To resolve the scale ambiguity and place the person in the correct global position, we jointly optimize the human and camera motion to minimize the discrepancy between the observed evidence and the estimated motion, while maintaining the plausibility of the human motion with a diffusion prior through the proposed control-inpainting SDS.

\subsubsection{Motion Representation.}
The camera motion is represented by the trajectory $\mathcal{C} = \{ (\mathbf{R}^{(i)}, \mathbf{t}^{(i)}) \}_{i=1}^{T}$, where $[\mathbf{R}^{(i)}, \mathbf{t}^{(i)}]$ is the camera pose at the $i$-th frame, consisting of the rotation matrix $\mathbf{R}^{(i)} \in \mathbb{R}^{3\times3}$ and the translation vector $\mathbf{t}^{(i)} \in \mathbb{R}^3$. The human motion is represented by the human trajectory $\mathbf{H} = \{\mathbf{h}^{(i)}\}_{i=1}^T$, where $\mathbf{h}^{(i)}=[\tau^{(i)}, \Phi^{(i)}, \theta^{(i)}, f^{(i)}, \beta]$ is the human pose at the $i$-th frame, consisting of the global translation $\tau^{(i)} \in \mathbb{R}^3$, global orientation $\Phi^{(i)} \in \mathbb{R}^3$, body pose parameters $\theta^{(i)} \in \mathbb{R}^{23\times3}$, foot contact labels $f \in \{0, 1\}^4$, and the body shape parameters $\beta \in \mathbb{R}^{10}$. We use the SMPL model~\cite{SMPL:2015} to represent the human pose and shape.
Before introducing our approach, we first revisit the formulation and drawbacks of SDS in Sec.~\ref{sec:revisit_sds}. Then we introduce the proposed control-inpainting SDS in Sec.~\ref{sec:coin_sds}. Finally, we present the global optimization pipeline with the proposed human-scene interaction loss in Sec.~\ref{sec:opt}.

\subsection{Revisiting SDS}
\label{sec:revisit_sds}

Score Distillation Sampling (SDS) was first introduced to distill 3D assets from pre-trained 2D text-to-image diffusion models~\cite{poole2023dreamfusion}. It exploits the knowledge from the diffusion models by seeking modes for the conditional distribution in the DDPM latent space to optimize the 3D scene representation. Similarly, we can optimize global human motion by distilling knowledge from a pre-trained motion diffusion model.

Given an global human motion $\mathbf{H}$, the marginal distribution of noisy latent $\mathbf{H}_t$ at timestep $t \in \mathcal{U}(0, 1)$ is defined as:
\begin{equation}
    q(\mathbf{H}_t | \mathbf{H}) = \mathcal{N}(\mathbf{H}_t; \sqrt{\bar{\alpha}_t} \mathbf{H}, (1 - \bar{\alpha}_t)\mathbf{I}),
\end{equation}
where $\bar{\alpha}_t \in (0, 1)$ is a hyperparameter controlled by the variance schedule of the diffusion model. SDS adopts the pre-trained diffusion model $\mathcal{D}_\phi({\mathbf{H}}_t, t, y)$, which takes in $\mathbf{H}_t$ and is used to model the conditional density of the human motion, where $\phi$ are the parameters of the diffusion model and $y$ is the condition. Then, SDS aims to distill global human motion $\mathbf{H}$ via seeking modes of the learned conditional density,
which can be achieved by a weighted denoising score matching objective:
\begin{equation}
    \min_{\mathbf{H}} \mathcal{L}_{\text{SDS}} := \mathbb{E}_{t,\boldsymbol{\epsilon}} \left[ \omega(t) \| \boldsymbol{\epsilon}^t_\phi - \boldsymbol{\epsilon} \|^2_2 \right],
\label{eq:sds_eps}
\end{equation}
where $\boldsymbol{\epsilon}^t_\phi$ is the predicted denoising direction from the diffusion model, $\mathbf{H}_t \sim q(\mathbf{H}_t | \mathbf{H})$ is sampled using the reparameterization trick, $\boldsymbol{\epsilon}$ is the corresponding sampled noise, and $\omega(t)$ is a weighting function that depends on the timestep $t$.

To clearly review the effect of SDS, we can reparameterize Eq.~\ref{eq:sds_eps} as:
\begin{equation}
    \min_{\mathbf{H}} \mathcal{L}_{\text{SDS}} := \mathbb{E}_{t} \left[ \frac{\omega(t)\sqrt{\bar{\alpha}_t}}{\sqrt{1 - \bar{\alpha}_t}} \left\| \mathbf{H} - \hat{\mathbf{H}}^t_0 \right\|^2_2 \right],
\label{eq:sds_h0}
\end{equation}
where
\begin{equation}
\hat{\mathbf{H}}^t_0 = \frac{\mathbf{H}_t - \sqrt{1 - \bar{\alpha}_t}\boldsymbol{\epsilon}^t_\phi}{\sqrt{\bar{\alpha}_t}}.
\end{equation}
Based on this reparameterization, we can see that the SDS objective is to minimize the discrepancy between global human motion $\mathbf{H}$ and the denoised global human motion $\hat{\mathbf{H}}^t_0$ from the motion diffusion model in a single step. The denoised motion $\hat{\mathbf{H}}^t_0$ serves as the \textit{pseudo ground truth}. However, at each optimization step, we randomly sample $t$ and $\boldsymbol{\epsilon}$ to generate the noisy latent ${\mathbf{H}}_t$, and we found the pre-trained diffusion model is sensitive to the input. Minor fluctuations in the input latent would substantially change the denoised motion, which leads to inconsistency in $\hat{\mathbf{H}}^t_0$ across different time steps.

Although randomness can help generate diverse plausible motions to infer occluded regions and unknown information, we do not need it for well-observed regions, such as simple body poses in a clean background. Such randomness in the denoising steps makes the generated $\hat{\mathbf{H}}^t_0$ difficult to align with the local 2D observations and results in wrong global human motion.
Moreover, this pseudo ground truth $\hat{\mathbf{H}}^t_0$ is generated from only a single denoising step, where the diffusion models may not produce high-quality motions, resulting in foot sliding and floating. 
Although sampling with a smaller timestep $t$ can alleviate these issues, the initial motion 
 is usually inaccurate and the denoiser is not able to remove artifacts with a small $t$. To exploit the knowledge of the motion diffusion model and denoise the initial motion, we must allow the SDS to sample with a larger timestep $t$ while maintaining high quality, consistency, and alignment with the local 2D observations.

\subsection{Control-Inpainting SDS}
\label{sec:coin_sds}
The limitations of SDS originate from the randomness and inconsistency of the denoised motion $\hat{\mathbf{H}}_0^t$, which serves as the pseudo ground truth in the objective function. 
To overcome this issue, we propose a novel \textbf{CO}ntrol-\textbf{IN}painting SDS (COIN-SDS) to generate \textit{high-quality} and \textit{consistent} pseudo-ground-truth motions. Our solution has three key ingredients (shown in Alg.~\ref{alg:coin}). First, to achieve \textit{high-quality} motions, we seek to produce the pseudo ground truth with multiple DDIM denoising steps. Second, to encourage \textit{consistent} motions, we propose to use partially observed evidence from the video as a control signal to dynamically guide the diffusion model and align the generated motions with the observations. Third, to further align the motion with observed regions, we propose a soft inpainting strategy within the denoising process.

\subsubsection{Multiple Denoising Steps.}
Intuitively, to obtain high-quality pseudo ground truth for SDS, we can replace the single-step denoised motion $\hat{\mathbf{H}}^t_0$ with a multi-step one $\tilde{\mathbf{H}}^t_0 := \tilde{\mathbf{H}}_0$, following the multi-step DDIM denoising process~\cite{song2020denoising}:
\begin{equation}
    \tilde{\mathbf{H}}_{t - \Delta t} = \sqrt{\bar{\alpha}_{t - \Delta t}}\cdot\hat{\mathbf{H}}_0^t + \sqrt{1 - \bar{\alpha}_{t - \Delta t}}\cdot\boldsymbol{\epsilon}^t_\phi,
\end{equation}
until $\tilde{\mathbf{H}}_0 = \tilde{\mathbf{H}}_{t - \Delta t}$ is obtained. By replacing $\hat{\mathbf{H}}_0^t$ in Eq.~\ref{eq:sds_h0} with $\tilde{\mathbf{H}}_0$, we can obtain a new objective for SDS:
\begin{equation}
    \min_{\mathbf{H}} \mathcal{L}_{\text{SDS}} := \mathbb{E}_{t} \left[ \frac{\omega(t)\sqrt{\bar{\alpha}_t}}{\sqrt{1 - \bar{\alpha}_t}} \left\| \mathbf{H} - \tilde{\mathbf{H}}_0 \right\|^2_2 \right].
    \label{eq:ddim}
\end{equation}
Although the multi-step denoising process can produce high-quality pseudo ground truth, it is computationally expensive to perform multiple denoising steps during optimization, which limits the practicality of increasing the number of denoising steps. In our experiments, we find that using $10$ denoising steps is sufficient to produce high-quality pseudo ground truth.

\begin{algorithm}[t]
    \caption{COIN-SDS}
    \label{alg:coin}
    \KwIn{Latest human motion ${\mathbf{H}}$, confidence score $\mathbf{S}$, visible mask $\mathbf{M}$}
    \KwOut{$\mathcal{L}_{\text{COIN-SDS}}$}

    Sample: $\tilde{\mathbf{H}}_{t} \sim \mathcal{N}(\mathbf{H}_{t}; \sqrt{\bar{\alpha}_{t}} {\mathbf{H}}, (1 - \bar{\alpha}_{t})\mathbf{I})$, $t \sim \mathcal{U}(0, 1)$\; 
    \For(\tcp*[f]{multi-step DDIM denoising}){$\bar{t} = [t, t - \Delta t, \ldots, \Delta t]$}
    {
    $\tilde{\mathbf{H}}_0^{\bar{t}, \text{known}} \leftarrow {\mathbf{H}}$\;
    $\tilde{\mathbf{H}}_0^{\bar{t}, \text{unknown}} \leftarrow \mathcal{D}_{\phi, \phi_c}(\tilde{\mathbf{H}}_{\bar{t}}, \bar{t}, {\mathbf{H}} \odot \mathbf{M})$ \tcp*{controlled denoising}
    $\tilde{\mathbf{M}} \leftarrow w(\bar{t}) * \mathbf{S} \odot \mathbf{M}$\;
    $\tilde{\mathbf{H}}^{\bar{t}}_0 \leftarrow \tilde{\mathbf{M}} \odot \tilde{\mathbf{H}}_0^{\bar{t}, \text{known}} + (1 - \tilde{\mathbf{M}}) \odot \tilde{\mathbf{H}}_0^{\bar{t}, \text{unknown}}$ \tcp*{soft inpainting}
    $\boldsymbol{\epsilon}^{\bar{t}}_\phi \leftarrow \frac{\tilde{\mathbf{H}}_{\bar{t}} - \sqrt{\bar{\alpha}_{\bar{t}}}\tilde{\mathbf{H}}^{\bar{t}}_0}{\sqrt{1 - \bar{\alpha}_{\bar{t}}}}$\;
    $\tilde{\mathbf{H}}_{\bar{t}-\Delta t} \leftarrow \sqrt{\bar{\alpha}_{\bar{t} - \Delta t}}\cdot\tilde{\mathbf{H}}^{\bar{t}}_0 + \sqrt{1 - \bar{\alpha}_{\bar{t} - \Delta t}}\cdot\boldsymbol{\epsilon}^{\bar{t}}_\phi$ \tcp*{update latent motion}
    }
    $\mathcal{L}_{\text{COIN-SDS}} = \frac{\omega(t)\sqrt{\bar{\alpha}_{t}}}{\sqrt{1 - \bar{\alpha}_{t}}} \left\| {\mathbf{H}} - \tilde{\mathbf{H}}_0 \right\|^2_2 $
\end{algorithm}

\subsubsection{Dynamic Controlled Sampling.}
To generate consistent motions that are aligned with the observed evidence, we propose to attach a control branch $\phi_c$ to the pre-trained diffusion model $\mathcal{D}_\phi$ to guide the motion generation. Given a latent motion $\tilde{\mathbf{H}}_t$, control signal ${\mathbf{c}}$, and the visible mask $\mathbf{M}$, we train a controlled denoiser $\mathcal{D}_{\phi, \phi_c}$ to generate intermediate denoised motion $\tilde{\mathbf{H}}^{t}_0$ for DDIM denoising:
\begin{equation}
    \tilde{\mathbf{H}}^{t}_0 = \mathcal{D}_{\phi, \phi_c}(\tilde{\mathbf{H}}_t, t, \mathbf{c} \odot \mathbf{M}),
    \label{eq:controlled_denoiser}
\end{equation}
where $\mathbf{c}$ and $\mathbf{M}$ are the same size as the motion, $\mathbf{M}$ is a binary mask with ones in observed pose dimensions, and $\odot$ denotes the element-wise multiplication. During training, we synthesize noise and occlusions by randomly adding Gaussian noise to the control signal $\mathbf{c}$ and randomly masking the pose and trajectory dimensions in $\mathbf{M}$. Details are provided in the appendix.

When using the denoiser in the optimization stage, instead of always using the initial noisy estimation from HybrIK~\cite{li2020hybrik} as the fixed control signal, we propose to use a \textit{dynamic control strategy}. Specifically, we use the optimized human motion from the previous iteration as the control signal, \ie, $\mathbf{c} = {\mathbf{H}}$. This strategy prevents performance degradation due to inaccurate initializations. A better control signal can guide the model to generate a more plausible pseudo motion, which in turn helps to optimize the global human motion and provides a control signal that better aligns with the input videos. Such self-evolving control signals can help to generate well-aligned global human motions.

The pre-trained motion diffusion model adopts a transformer encoder structure like \cite{tevet2022human}. We follow ControlNet~\cite{zhang2023adding} to encode the control signals and guide the denoiser output. We create a trainable copy of 4 encoding blocks of the pre-trained motion diffusion model followed with zero convolutions. The input to the control branch is the concatenation of the latent and the control signals. The AMASS~\cite{AMASS:ICCV:2019} dataset is used to train the motion diffusion model. The finetuning of the controlled denoiser is computationally efficient since the pre-trained branch is frozen and only the control branch is trained. See the appendix for more details on the architecture and training settings.

\subsubsection{Soft Inpainting.}

While guiding motion generation encourages outputs to align with the conditions, it is often not strong enough. We further seek to improve the consistency by masking the known regions and inpainting the unknown regions.
Given a binary mask $\mathbf{M}$ that indicates the observed and unobserved regions, we can use the diffusion model to generate the inpainted motion $\tilde{\mathbf{H}}_0^t$ following the DDIM denoising process:
\begin{subequations}
    \begin{align}
    &\tilde{\mathbf{H}}_0^{t, \text{known}} = {\mathbf{H}}, \\
    &\tilde{\mathbf{H}}_0^{t, \text{unknown}} = \mathcal{D}_{\phi, \phi_c}(\tilde{\mathbf{H}}_t, t, {\mathbf{H}} \odot \mathbf{M}),\\
    &\tilde{\mathbf{H}}^{t}_0 = \mathbf{M} \odot \tilde{\mathbf{H}}_0^{t, \text{known}} + (1 - \mathbf{M}) \odot \tilde{\mathbf{H}}_0^{t, \text{unknown}}.
    \end{align}
\end{subequations}
Thus, the known regions are overwritten with the observations, while the unknown regions are sampled from the diffusion model. However, the above formulation keeps the observed parts unchanged during the denoising process. In practice, the observed parts can be noisy and not perfect. We still want the diffusion model to refine the observed parts but not change them significantly.

Here, we present a soft inpainting strategy to infill the unobserved regions while refining the observed regions by dynamically reweighting the denoised direction from the diffusion model. Specifically, instead of using a binary mask, we adopt a continuous mask $\tilde{\mathbf{M}}$ depending on both the confidence score of the observations $\mathbf{S}$ and the denoising time step $t$:
\begin{equation}
    \tilde{\mathbf{M}} = w(t) * \mathbf{S} \odot \mathbf{M},
\end{equation}
where we set $w(t) = \max(0, \frac{t - 0.5}{0.5})$ to linearly decrease the weight of the observations as the denoising time step decreases. As the time step decreases, the denoising process will be more deterministic and the model will be more certain of the generated motions.

Combining the three components of our solution, the final objective for COIN-SDS is formulated as:
\begin{equation}
    \min_{\mathbf{H}} \mathcal{L}_{\text{COIN-SDS}} := \mathbb{E}_{t} \left[ \frac{\omega(t)\sqrt{\bar{\alpha}_t}}{\sqrt{1 - \bar{\alpha}_t}} \left\| \mathbf{H} - \tilde{\mathbf{H}}_0(\mathbf{H}, \mathbf{M}, \mathbf{S}, t) \right\|^2_2 \right].
\end{equation}
We summarize \acronym in Alg.~\ref{alg:coin}.

\subsection{Global Optimization}
\label{sec:opt}

Here we present the overall optimization pipeline for the joint estimation of global human and camera motion with the proposed COIN-SDS loss. Note that we use SLAM to initialize the camera poses, which is scale-ambiguous. Therefore, we need to jointly optimize the camera scale $s$ with the human and camera motions. Furthermore, the SLAM method assumes the camera in the first frame to be at the origin. To put the human motion in the correct positions, we follow PACE~\cite{kocabas2024pace} and also optimize the camera height $h_0$ and the orientation $R_0$ for the first frame. The global human motions are initialized by the estimated local motions and the camera poses. The overall optimization objective is:
\begin{equation}
    \min_{\mathbf{H}, \mathcal{C}, s, h_0, R_0, \beta} \mathcal{L}_{\text{body}} + \mathcal{L}_{\text{COIN-SDS}} + \mathcal{L}_{\text{HSR}},
\end{equation}
where
\begin{equation}
    \mathcal{L}_{\text{body}} = \mathcal{L}_{\text{2D}} + \mathcal{L}_{\text{3D}} + \mathcal{L}_{\beta} + \mathcal{L}_{\text{smooth}} + \mathcal{L}_{\text{contact}}.
\end{equation}
$\mathcal{L}_{\text{2D}}$ measures the 2D reprojection error between the projected 2D body joints of the estimated human motion and the detected 2D keypoints from an off-the-shelf 2D joint detector~\cite{mmpose2020}. $\mathcal{L}_{\text{3D}}$ measures the distance between the estimated local 3D joints and the detected 3D joints from an off-the-shelf 3D joint detector~\cite{Sarandi2023dozens}. $\mathcal{L}_{\beta}$ is the shape regularization loss. $\mathcal{L}_{\text{smooth}}$ is the temporal smoothness loss. $\mathcal{L}_{\text{contact}}$ is the foot contact loss to encourage zero velocities for contact joints. The contact labels are obtained from the pseudo ground truth motion from COIN-SDS. Please refer to the appendix for more details.

\subsubsection{Human-scene Relation Loss.} The camera trajectories recovered from SLAM are scale-ambiguous. Previous works~\cite{ye2023slahmr,kocabas2024pace} optimize the camera scale by projecting the global human motion to the camera space using the camera poses and minimizing the reprojection error. However, such a method entirely relies on the global human motions, which is in turn affected by the camera scale. If the human motion is not initialized well, the camera scale will also be inaccurate. To solve this problem, instead of the global human motions, we propose a new human-scene relation loss that uses the depth relation between the human and scene in the camera space, which disentangles the effect of the camera itself. 

Specifically, we use the point cloud of the scene recovered by SLAM as a constraint. First of all, the scale of the scene point cloud is the same as the camera scale, so optimizing the scene scale is equivalent to optimizing the camera scale. Second, the scene points that are projected onto the visible vertices of the body mesh should be occluded by the person. Otherwise, the corresponding body parts are invisible. Therefore, we can constrain the depth of the occluded scene points to be larger than the depth of the human body vertices. While finding the corresponding body vertices for each scene point is time-consuming, we propose to use the depth of its nearest body joint as a proxy. Given the scene point cloud $\mathcal{P}$ and the camera scale $s$, the human-scene relation loss is formulated as:
\begin{equation}
    \mathcal{L}_{\text{HSR}} = - \frac{1}{|\mathcal{P}|} \sum_{i=1}^T \sum_{{p} \in \mathcal{P}^*} \min(0, \mathcal{T}^{(i)}(p)_z - j^{(i)}(p)_z )\cdot\mathds{1}(\mathcal{T}^{(i)}(p) \text{ is invisible}),
\end{equation}
where $\mathcal{P}^* = \mathcal{P} * s$ is the scaled point cloud of the scene, $\mathcal{T}^{(i)}(p) = \mathbf{R}^{(i)}p + \mathbf{t}^{(i)}$ is the transformed point in the $i$-th frame, $j^{(i)}(p)$ is body joint that has the nearest 2D projection to the scene point $p$ in the $i$-th frame, and $z$ denotes the depth of a given point. If the depth order is correct, \ie, $\mathcal{T}^{(i)}(p)_z - j^{(i)}(p)_z > 0$, the loss is zero. The proposed human-scene relation loss uses the relation between the local motions and the scene to alleviate the scale ambiguity. The depth relation regularizes consistency among humans, cameras, and scenes.

\section{Experiments}
\label{sec:exp}

\subsubsection{Datasets.}
We perform experiments on three human motion datasets. First is the real-world dataset RICH~\cite{huang2022cap}. We follow previous works~\cite{yuan2022glamr, kocabas2024pace, shin2024wham} to assess the performance of global human motion estimation using this dataset. The second one is EMDB~\cite{kaufmann2023emdb}. We follow previous works~\cite{shin2024wham} to evaluate on a subset of EMDB for which they provide ground truth global motion with dynamic cameras. The third dataset is HCM~\cite{kocabas2024pace}, a synthetic dataset. Compared to real-world datasets, HCM contains more challenging camera motions. We follow previous work~\cite{kocabas2024pace} to evaluate the global human motion and the camera motion using this dataset.

\subsubsection{Metrics.} We report various metrics for both human and camera motion. For human motion, standard metrics W-MPJPE and WA-MPJPE are used to evaluate global motion, while PA-MPJPE evaluates local motion. We also include an ACCEL metric to measure the joint acceleration error. For evaluation on EMDB, we follow previous work~\cite{shin2024wham} to split sequences into smaller chunks of $100$ frames and align each output segment with the ground-truth data using the first two frames W-MPJPE$_{100}$ or the entire segment WA-MPJPE$_{100}$. Root Orientation Error (ROE in $\mathtt{deg}$) and Root Translation Error (RTE in $m$) evaluate the error over the entire trajectory after aligning with the initial camera pose.

For camera motion, we report the average translation error after scale alignment (ATE), without scale alignment (ATE-S), and the camera acceleration error (CAM ACCEL). ATE-S more accurately reflects inaccuracies in the captured scale of the camera.

\subsubsection{Baselines.} As discussed in Sec.~\ref{sec:related}, there are different ways to use the pre-trained motion diffusion model as a motion prior. Here, we summarize the three main solutions and compare them with COIN in Sec.~\ref{sec:ablation}. (\textbf{1}) Guided Sampling, which embeds analytical guidance within the denoising procedure using objective functions, such as 2D projection and foot contact consistency. This does not suit our task because the camera trajectories are also unknown and guided gradients from the wrong camera will lead to unrealistic human motions. We need to optimize the human motion and camera motion simultaneously. (\textbf{2}) Noise Optimization, which represents the motion as latent noise and directly optimizes it. This is similar to other motion priors such as VAE~\cite{he2022nemf}. At each optimization step, we need to calculate the gradients of the latent w.r.t. the generated motion, which is computationally expensive and we find the performance is not good enough. (\textbf{3}) Vanilla SDS, which removes our design and directly uses SDS for optimization.

\begin{table}[t]
    \begin{center}
        \caption{\textbf{Global human motion estimation on the RICH dataset.}}
        \vspace{-6mm}
        \label{table:rich}
        \resizebox{\textwidth}{!}{
            
        \begin{tabular}{l|ccccc}

        \toprule
        Method & ~PA-MPJPE~$\downarrow$~ & ~W-MPJPE~$\downarrow$~ & ~WA-MPJPE~$\downarrow$~ & ~W-RJE~$\downarrow$~ & ~ACCEL~$\downarrow$~ \\
        \midrule
        HybrIK~\cite{li2020hybrik} + SLAM~\cite{teed2021droid} & 46.7 & 1073.1 & 404.4 & 1166.2 & 20.2 \\
        GLAMR~\cite{yuan2022glamr} & 79.9 & 653.7 & 365.1 & 646.6 & 107.7 \\
        SLAHMR~\cite{ye2023slahmr} & 52.5 & 571.6 & 323.7 & 400.5 & 9.4 \\
        WHAM~\cite{shin2024wham} & 46.2 & 497.6 & 272.7 & 478.2 & \textbf{6.7} \\
        PACE~\cite{kocabas2024pace} & 49.3 & 380.0 & 197.2 & 370.8 & 8.8 \\
        \midrule
        Guided Sampling & 132.8 & 1384.6 & 502.2 & 1440.9 & 24.2 \\
        Noise Optimization & 66.7 & 414.8 & 195.3 & 429.2 & 8.4 \\
        Vanilla SDS & 78.8 & 1453.5 & 497.2 & 1458.0 & 12.7 \\
        \midrule
        COIN w/o Controlled Sampling~ & 44.0 & 825.0 & 291.8 & 848.5 & 10.8 \\
        COIN w/o Dynamic Control & 49.5 & 293.8 & 180.6 & 299.1 & 8.6 \\
        COIN w/o Soft Inpainting & 47.6 & 325.8 & 196.0 & 324.9 & 9.6 \\
        COIN w/o $\mathcal{L}_{\text{HSR}}$ & 43.6 & 273.0 & 176.3 & 281.1 & 8.2 \\
        \midrule
        COIN & \textbf{42.9} & \textbf{254.5} & \textbf{169.5} & \textbf{249.9} & {7.5} \\

        \bottomrule
        \end{tabular}
        }
    \end{center}
\end{table}

\subsection{Comparison with State-of-the-Art Methods}

\subsubsection{Human Motion Estimation.} We compare COIN against state-of-the-art methods on the RICH, EMDB, and HCM datasets. Quantitative results are shown in Tabs.~\ref{table:rich},~\ref{table:emdb}, and~\ref{table:hcm_human}. We observe that COIN significantly outperforms the state-of-the-art methods on all datasets. On the RICH dataset, COIN outperforms the state-of-the-art method, PACE~\cite{kocabas2024pace}, by \textbf{125.5} mm in terms of W-MPJPE, showing \textbf{33.0}\% relative improvement. On the EMDB and HCM datasets, COIN shows \textbf{29.1} mm and \textbf{109.0} mm improvement in terms of W-MPJPE, respectively.
Qualitative comparisons with state-of-the-art methods, PACE~\cite{kocabas2024pace} and WHAM~\cite{shin2024wham}, are shown in \cref{fig:teaser,fig:qual}. More qualitative results are shown in the supplementary video.

COIN not only improves global motion, but also improves local body motion. In terms of PA-MPJPE, COIN shows \textbf{3.3} mm, \textbf{9.2} mm, and \textbf{19.8} mm improvement on the RICH, EMDB, and HCM datasets, respectively. This demonstrates that COIN is able to distill high-quality motion priors from the diffusion model and help both local and global motion estimation. Regarding the joint acceleration error, COIN is 0.8 mm/s$^2$ higher than WHAM~\cite{shin2024wham}. Note that the joint acceleration error reflects the smoothness. Human motion can be over-smoothed but not accurate, hence it is important to look at acceleration error in conjunction with other metrics.

\begin{table}[t]
    \begin{center}
        \caption{\textbf{Global human motion estimation on the EMDB dataset.}}
        \vspace{-6mm}
        \label{table:emdb}
        \resizebox{\textwidth}{!}
        {
            
        \begin{tabular}{l|ccccc}

        \toprule
        Method & ~PA-MPJPE~$\downarrow$~ & ~W-MPJPE$_{100}$~$\downarrow$~ & ~WA-MPJPE$_{100}$~$\downarrow$~ & ~RTE~$\downarrow$~ & ~ROE~$\downarrow$~ \\ 
        \midrule
        HMR2.0~\cite{goel2023humans} $+$ DPVO~\cite{teed2023deep}~ & 49.6 & 2320.9 & 662.9 & 17.5 & 44.4 \\ 
        GLAMR~\cite{yuan2022glamr} & 56.0 & 756.1 & 286.2 & 16.7 & 74.9 \\ 
        TRACE~\cite{sun2023trace} & 58.0 & 2244.9 & 544.1 & 18.9 & 72.7 \\ 
        SLAHMR~\cite{ye2023slahmr} & 61.5 & 807.4 & 336.9 & 13.8 & 67.9 \\ 
        WHAM~\cite{shin2024wham} & 41.9 & 439.2 & 166.1 & 8.4 & 36.3 \\ 
        \midrule
        COIN & \textbf{32.7} & \textbf{407.3} & \textbf{152.8} & \textbf{3.5} & \textbf{34.1} \\

        \bottomrule
        \end{tabular}
        }
    \end{center}
\end{table}

\begin{table}[t]
    \begin{center}
        \caption{\textbf{Global human motion estimation on the HCM dataset.}}
        \vspace{-6mm}
        \label{table:hcm_human}
        \resizebox{\textwidth}{!}{
            
        \begin{tabular}{l|ccccc}

        \toprule
        Method & ~PA-MPJPE~$\downarrow$~ & ~W-MPJPE~$\downarrow$~ & ~WA-MPJPE~$\downarrow$~ & ~W-RJE~$\downarrow$~ & ~ACCEL~$\downarrow$~ \\
        \midrule
        HybrIK~\cite{li2020hybrik} + SLAM~\cite{teed2021droid}~ & 67.6 & 1137.3 & 780.3 & 1100.9 & 51.3 \\
        GLAMR~\cite{yuan2022glamr} & 86.0 & 1977.6 & 653.8 & 1958.0 & 33.4 \\
        SLAHMR~\cite{ye2023slahmr} & 69.9 & 888.9 & 483.5 & 862.2 & 14.9 \\
        PACE~\cite{kocabas2024pace} & 65.3 & 861.2 & 478.3 & 839.5 & 16.7 \\
        WHAM~\cite{shin2024wham} & 47.9 & 588.9 & 279.3 & 579.2 & 13.1 \\
        \midrule
        COIN & \textbf{45.5} & \textbf{479.9} & \textbf{212.1} & \textbf{470.7} & \textbf{10.1} \\

        \bottomrule
        \end{tabular}
        }
    \end{center}
\end{table}

\subsubsection{Camera Motion Estimation.} We further evaluate the performance of COIN on camera motion estimation on the HCM dataset. Quantitative results are shown in Tab.~\ref{table:hcm_cam}. COIN substantially surpasses the state-of-the-art camera motion estimation methods. Specifically, COIN reduces the absolute camera translation error, ATE-S by \textbf{73.8} mm. This demonstrates that COIN is able to disentangle human and camera motions and accurately estimate the camera motion.

\begin{table}[t]
    \begin{center}
        \caption{\textbf{Camera motion estimation on the HCM dataset.}}
        \vspace{-3mm}
        \label{table:hcm_cam}
        {
            
        \begin{tabular}{l|ccc}

        \toprule
        Method & ~ATE~$\downarrow$~ & ~ATE-S~$\downarrow$~ & ~CAM ACCEL~$\downarrow$~ \\
        \midrule
        HybrIK~\cite{li2020hybrik} + SLAM~\cite{teed2021droid} & 155.8 & 1670.7 & 17.1 \\
        GLAMR~\cite{yuan2022glamr} & 1295.2 & 1714.6 & 282.9 \\
        SLAHMR~\cite{ye2023slahmr} & 155.8 & 506.5 & 17.6 \\
        PACE~\cite{kocabas2024pace} & 137.5 & 459.7 & 16.2 \\
        \midrule
        Guided Sampling & 335.6 & 992.4 & 15.6 \\
        Noise Optimization & 206.0 & 500.9 & 12.3 \\
        Vanilla SDS & 306.4 & 656.4 & 13.7 \\
        \midrule
        COIN w/o Controlled Sampling~ & 299.6 & 553.8 & 14.0 \\
        COIN w/o Dynamic Control & 149.8 & 397.7 & \textbf{11.3}  \\
        COIN w/o Soft Inpainting & 167.7 & 423.8 & 11.4 \\
        COIN w/o $\mathcal{L}_{\text{HSR}}$ & 147.8 & 402.1 & 12.0 \\
        \midrule
        COIN & \textbf{135.3} & \textbf{385.9} & \textbf{11.3} \\
        \bottomrule
        \end{tabular}
        }
    \end{center}
\end{table}

\subsection{Ablation Study}
\label{sec:ablation}
In this section, we conduct ablation studies on the RICH and HCM datasets to evaluate the impact of each component on human and camera motions, respectively. More comparisons on other datasets are provided in the appendix.

\subsubsection{Baselines with Motion Diffusion Model.}
We first compare the aforementioned baselines with COIN. To use guided sampling in our tasks, we jointly update the camera poses using the gradients from the objective function during denoising. Quantitative results of human and camera motion estimation are shown in \cref{table:rich,table:hcm_cam}, respectively. We observe that COIN outperforms all the baselines in terms of all metrics. As expected, guided sampling shows a terrible performance because it is not able to accurately estimate the camera trajectories. Noise optimization is better than vanilla SDS but worse than COIN, which generates consistent motion priors. Vanilla SDS is replacing $\mathcal{L}_{\text{COIN}}$ with the SDS loss and keeping the rest of the settings the same. These results demonstrate the effectiveness of COIN over other motion diffusion baselines.

\subsubsection{Impact of Dynamic Controlled Sampling.} To study the effectiveness of controlled sampling, we compare COIN with and without using the controlled denoiser. Quantitative results are shown in \cref{table:rich,table:hcm_cam}. We observe that controlled sampling significantly improves the performance of COIN. Specifically, COIN with controlled sampling reduces the W-MPJPE and ATE-S by \textbf{570.5} mm and \textbf{167.9} mm, showing \textbf{69.3}\% and \textbf{30.3}\% relative improvement, respectively. This demonstrates that controlled sampling is able to generate high-quality motion priors that align with the observed evidence and help improve human and camera motion estimation.

\subsubsection{Impact of Soft Inpainting.} We further study the effectiveness of soft inpainting. Quantitative comparisons are shown in \cref{table:rich,table:hcm_cam}. We observe that while soft inpainting also improves W-MPJPE, it is much more effective than controlled sampling in terms of local body motions. Specifically, COIN with soft inpainting reduces the PA-MPJPE by \textbf{4.7} mm.

\begin{figure}[t]
    \centering
    \includegraphics[width=\linewidth]{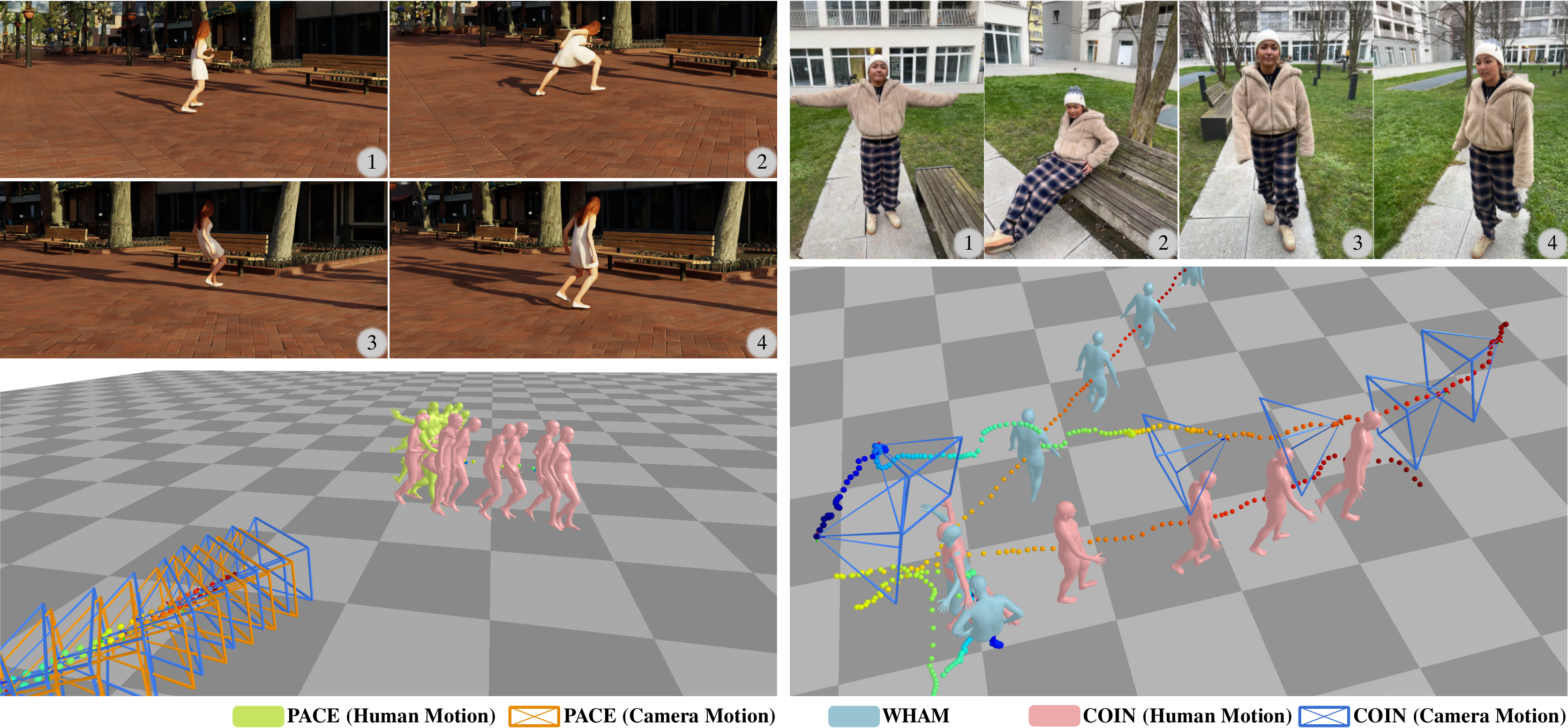}
    \vspace{-6mm}
    \caption{\textbf{Qualitative comparisons with state-of-the-art methods.} PACE~\cite{kocabas2024pace} fails to recover a correct trajectory (left). WHAM~\cite{shin2024wham} estimates the wrong walking direction of the person (right). Our approach, COIN, recovers the human and camera motion accurately in both scenarios.}
    \vspace{2mm}
    \label{fig:qual}
\end{figure}

\section{Conclusion}
\label{sec:conclusion}

In this paper, we propose COIN, a diffusion-based optimization framework for global human and camera motion estimation from dynamic cameras. We identify the inconsistency problem of distilling knowledge from the diffusion model with conventional SDS loss. To address this issue, COIN uses a controlled denoiser combined with soft inpainting to distill a high-quality, well-aligned, and consistent motion prior. To further address the scale ambiguity of the camera trajectory, we develop a novel human-scene relation loss that imposes consistency among the human motion, camera motion, and scene features. Comprehensive experiments on challenging synthetic and real-world datasets demonstrate the effectiveness of COIN, which outperforms the SOTA by a large margin in recovering accurate global human motion and camera motion.

\textbf{Limitations and Future Works:} While COIN is able to jointly optimize camera trajectories and global human motions, it requires initialization from SLAM. If the SLAM method fails catastrophically, COIN may not be effective. Additionally, we found COIN fails under severe occlusions where there are several unseen frames and the diffusion model cannot provide consistent guidance. Another limitation is that COIN is an optimization framework, so it is unsuitable for real-time applications. Since the denoising diffusion models have shown their power to model data distributions in many different domains, looking forward we can learn the joint distribution of humans and cameras with the diffusion model to additionally denoise the artifacts in camera motions. Such a joint distribution model also holds the potential of real-time human and camera motion estimation via guided sampling with few DDIM denoising steps.

\appendix
\section*{Appendix}

\section{Controlled Denoiser}
\label{sec:arch}

\begin{figure}[h]
    \centering
    \includegraphics[width=\linewidth]{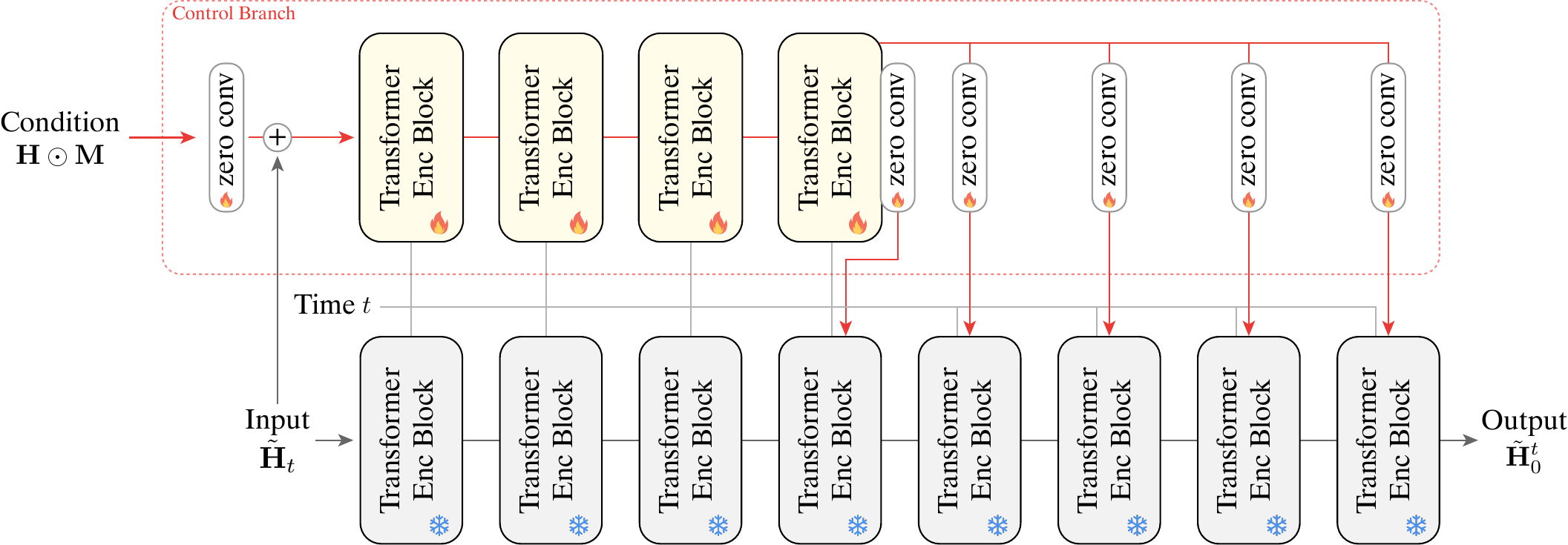}
    \caption{\textbf{Architecture of the controlled denoiser.}}
    \label{fig:arch}
\end{figure}

\subsubsection{Architecture Details.} The detailed architecture of the proposed controlled denoiser is illustrated in \cref{fig:arch}. We adopt a pre-trained transformer-based motion diffusion model as our backbone model. We create a trainable copy of the first $4$ encoder blocks. The condition is first encoded by a zero-initialized convolution layer and then concatenates with the input latent motion $\tilde{\mathbf{H}}_t$.
The outputs are followed by $5$ zero convolution layers and added to the last $5$ encoder blocks.

\subsubsection{Training Details.} We use the AMASS~\cite{AMASS:ICCV:2019} dataset to train the controlled denoiser. To simulate noisy motions in our application, we add Gaussian noise to the conditions. For the root orientation, the noise level is set to $0.05$. For the body pose, the noise level is set to $0.01$. For the translation, the noise level is set to $0.1$. To simulate occlusions, we randomly mask the conditions. With $0.5$ probability, all global trajectories are masked out; with $0.5$ probability, all global root orientations are masked out. The probabilities of the above two cases are calculated independently, \ie, it is possible to mask out the trajectories and orientations at the same time. With $0.2$ probability, the lower half of the body is masked out; with $0.2$ probability, the entire local pose is masked out; with $0.5$ probability, we randomly mask body joints, and each joint is masked with $0.3$ probability.

\section{Ablation Study}
\label{sec:supp_ablation}

\noindent\textbf{Impact of Each Component.} To comprehensively evaluate the impact of each component of COIN, we further conduct ablation studies on the EMDB~\cite{kaufmann2023emdb} and HCM~\cite{kocabas2024pace} datasets. Quantitative results are shown in \cref{table:supp_emdb,table:supp_hcm_human}. COIN shows consistent improvement against other diffusion-based baselines.


\begin{table}[h]
    \begin{center}
        \caption{\textbf{Global human motion estimation on the EMDB dataset.}}
        \label{table:supp_emdb}
        \resizebox{\textwidth}{!}
        {
            
        \begin{tabular}{l|ccccc}

        \toprule
        Method & ~PA-MPJPE~$\downarrow$~ & ~W-MPJPE$_{100}$~$\downarrow$~ & ~WA-MPJPE$_{100}$~$\downarrow$~ & ~RTE~$\downarrow$~ & ~ROE~$\downarrow$~ \\ 
        \midrule
        Noise Optimization & 53.9 & 873.8 & 275.8 & 10.4 & 96.4 \\
        Guided Sampling & 107.5 & 1713.9 & 462.8 & 7.2 & 71.5 \\
        Vanilla SDS & 64.5 & 1310.3 & 520.0 & 12.3 & 83.0 \\
        \midrule
        COIN w/o Controlled Sampling~ & 39.6 & 815.2 & 338.7 & 7.8 & 44.3 \\
        COIN w/o Dynamic Control & 36.4 & 441.2 & 162.1 & 4.1 & 40.2 \\
        COIN w/o Soft Inpainting & 35.1 & 495.4 & 176.8 & 4.8 & 43.6 \\
        COIN w/o $\mathcal{L}_{\text{HSR}}$ & 33.0 & 461.3 & 162.6 & 4.0 & 38.4 \\
        \midrule
        COIN & \textbf{32.7} & \textbf{407.3} & \textbf{152.8} & \textbf{3.5} & \textbf{34.1} \\

        \bottomrule
        \end{tabular}
        }
    \end{center}
\end{table}

To further evaluate the effect of each individual loss, we report the W-MPJPE on the RICH dataset.

\begin{table}[ht]
    \centering
    \resizebox{\linewidth}{!}
    {
    \begin{tabular}{c|c|c|c|c|c|c|c}
        COIN (full) & w/o $\mathcal{L}_{2D}$ & w/o $\mathcal{L}_{3D}$ &w/o $\mathcal{L}_{\beta}$ & w/o $\mathcal{L}_{smooth}$ & w/o $\mathcal{L}_{contact}$ & w/o $\mathcal{L}_{SDS}$ & w/o $\mathcal{L}_{HSR}$ \\
        \midrule
        254.5 & 448.7 & 329.4 & {279.9} & {256.1} & {270.3} & {480.6} & 273.0 \\
    \end{tabular}
    }
    \vspace{-5mm}
\end{table}

\begin{figure}[ht]
    \centering
    \includegraphics[width=\linewidth]{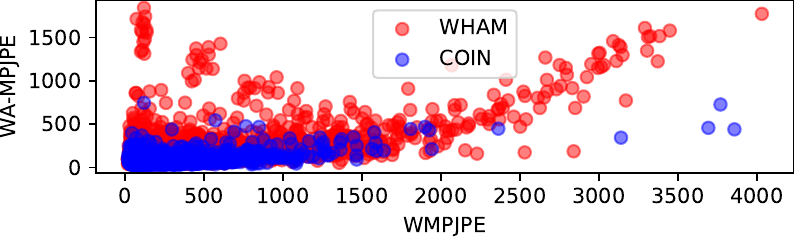}
    \caption{Error distributions on the EMDB dataset.}
    \label{fig:scatter}
\end{figure}

\noindent\textbf{Error Distribution.}
We further present the error distribution on the EMDB dataset to show more details of the COIN predictions. We also plot the error distribution of WHAM~\cite{shin2024wham} for comparison. The scatter plot is shown in \cref{fig:scatter}. Here we follow the evaluation protocol of EMDB and evaluate W-MPJPE and WA-MPJPE per $100$ frames. It is shown that COIN is more robust and has fewer outlier predictions.

\noindent\textbf{SLAM vs SfM.}
Compared to SLAM, SfM methods are stronger baselines for camera motion estimation. Because our cases always contain dynamic objects, we used ParticleSfM~\cite{zhao2022particlesfm} for its ability to handle dynamic objects and compared it to SLAM. Comparisons are shown in \cref{table:sfm}. Note that SfM methods run extremely slow compared to SLAM. Given a video with 700 frames on the RICH dataset~\cite{huang2022cap}, ParticleSfM takes over 17 hours while DROID-SLAM only needs 4 minutes. For videos on the EMDB dataset with more than 2000 frames, ParticleSfM took a few days to finish, and could not converge for many. Hence, SLAM is a more viable choice for in-the-wild videos. If we replace DROID-SLAM with ParticleSfM, on the converged videos, the baseline results improve by 200 mm. However, it has minimal impact on COIN demonstrating the robustness of our method to SLAM errors. We would like the emphasize that ParticleSfM could not converge on many of the EMDB videos and the results below are only the subset where it converged.

\begin{table}[h]
    \centering
    \caption{\textbf{SLAM vs. ParticleSfM on the converged subset of the EMDB dataset.}}
    \label{table:sfm}
    {
    \begin{tabular}{c|c|c|c|c}
        & HybrIK + SLAM & HybrIK + SfM & COIN (SLAM) & COIN (SfM) \\
        \midrule
        W-MPJPE & 643.9 & 439.0 & 350.1 & 330.9
    \end{tabular}
    }
\end{table}



\begin{table}[h]
    \begin{center}
        \caption{\textbf{Global human motion estimation on the HCM dataset.}}
        \label{table:supp_hcm_human}
        \resizebox{\textwidth}{!}{
            
        \begin{tabular}{l|ccccc}

        \toprule
        Method & ~PA-MPJPE~$\downarrow$~ & ~W-MPJPE~$\downarrow$~ & ~WA-MPJPE~$\downarrow$~ & ~W-RJE~$\downarrow$~ & ~ACCEL~$\downarrow$~ \\
        \midrule
        Noise Optimization & 66.0 & 813.9 & 328.9 & 794.7 & 10.2 \\
        Guided Sampling & 118.2 & 1653.0 & 635.4 & 1626.7 & 24.7 \\
        Vanilla SDS & 59.0 & 1108.2 & 569.2 & 1102.6 & 11.8 \\
        \midrule
        COIN w/o Controlled Sampling~ & 47.6 & 904.8 & 428.5 & 898.9 & 11.0 \\
        COIN w/o Dynamic Control & 47.4 & 486.9 & 239.7 & 477.5 & 10.2 \\
        COIN w/o Soft Inpainting & 48.6 & 487.1 & 264.1 & 478.0 & 10.8 \\
        COIN w/o $\mathcal{L}_{\text{HSR}}$ & 47.0 & 488.5 & 219.3 & 479.4 & 10.1 \\
        \midrule
        COIN & \textbf{45.5} & \textbf{479.9} & \textbf{212.1} & \textbf{470.7} & \textbf{10.1} \\

        \bottomrule
        \end{tabular}
        }
    \end{center}
\end{table}

\section{Global Optimization}
\label{sec:supp_opt}

Here we detail our optimization formulation for the reconstruction of global human and camera motion. Simultaneously optimizing both camera motion and global human motion can result in local minima. To address this challenge, we follow PACE~\cite{kocabas2024pace} and adopt a multi-stage optimization pipeline. Before running optimization, we initialize the global human motion with the noisy observations using the controlled denoiser. We randomly sample a Gaussian noise and run DDPM to generate the global motion. In stage 1, we optimize only the first frame camera parameters $(R_0, h_0)$, camera scale $s$, and the body shape $\beta$. In stage 2, we optimize the first frame camera parameters $(R_0, h_0)$, camera scale $s$, the body shape $\beta$, and the global human motion $\mathbf{H}$. In stage 3, we jointly optimize the full camera trajectory along with the global human motion. Given a long video, we split it into windows of $T=128$ frames. We use $16$ overlapping frames to help reduce discontinuities across windows. The mask $\mathbf{M}$ is defined by thresholding the confidence scores of the detected 2D keypoints. The threshold is $0.3$. Each stage is run for $500$ steps. The learning rates of the 3 stages are $0.01$, $0.01$, and $0.001$, respectively. We use the Adam solver for optimization. Implementation is in PyTorch.


%
%
\bibliographystyle{splncs04}
\bibliography{egbib}
\end{document}